\documentclass[conference,dvipsnames,png,border=10pt,tikz]{IEEEtran}
\IEEEoverridecommandlockouts
\usepackage[utf8]{inputenc}
\usepackage{svg}
\usepackage{cite}
\usepackage{amsmath,amssymb,amsfonts}
\usepackage{algorithmic}
\usepackage{graphicx}
\usepackage{textcomp}
\usepackage{hyperref}
\usepackage{xcolor}
\usepackage{pdfpages}
\usepackage{url}
\usepackage{tikz}
\usepackage[margin=1in]{geometry}
\usepackage{tabularx}
\usepackage{enumitem}
\usepackage{tikz}
\usepackage{amsthm}
\usetikzlibrary{positioning}
\theoremstyle{definition}

\setlist{nolistsep}

\definecolor{blue}{HTML}{008ED7}
\definecolor{mygray}{gray}{0.75}
\definecolor{lightBlue}{HTML}{e5f7ff}

\usepackage{mathtools}
\usetikzlibrary{shapes,arrows}

\usepackage{ulem}

\tikzstyle{decision} = [diamond, draw, fill=blue!20, 
    text width=4.5em, centered text badly, node distance=3cm, inner sep=0pt]
\tikzstyle{block} = [rectangle, draw, fill=blue!20, 
    text width=5em, text centered, rounded corners, minimum height=4em]
\tikzstyle{line} = [draw, -latex']
\tikzstyle{cloud} = [draw, ellipse,fill=red!20, node distance=3cm,
    minimum height=2em]

\newtheorem{example}{Example}[section]

\def\BibTeX{{\rm B\kern-.05em{\sc i\kern-.025em b}\kern-.08em
    T\kern-.1667em\lower.7ex\hbox{E}\kern-.125emX}}

\usetikzlibrary{shapes.geometric} 
\usetikzlibrary{arrows, backgrounds, calc, hobby, positioning}

\ExplSyntaxOn
\cs_if_exist:NF \prg_stepwise_function:nnnN { \cs_gset_eq:NN \prg_stepwise_function:nnnN \int_step_function:nnnN }
\cs_if_exist:NF \prg_stepwise_inline:nnnn { \cs_gset_eq:NN \prg_stepwise_inline:nnnn \int_step_inline:nnnn }
\ExplSyntaxOff

\tikzset{vertex style/.style={
    draw=#1,
    thick,
    fill=#1!10,
    text=black,
    ellipse,
    minimum width=2cm,
    minimum height=0.75cm,
    font=\small,
    outer sep=3pt,
  },
  text style/.style={
    sloped,
    text=black,
    font=\footnotesize,
    above
  }
}

\usepackage{algorithm2e}
\begin{document}

\title{ISA-bEL: Intelligent Search Algorithm based on Entity Linking \\
}

\author{\IEEEauthorblockN{Rubén González Sendino}
\IEEEauthorblockA{\textit{Digital Lab} \\
\textit{Vodafone}\\
Madrid, Spain \\
\href{https://orcid.org/
0000-0002-0283-6739}{ORCID}}

\and
\IEEEauthorblockN{Mónica Ortega}
\IEEEauthorblockA{\textit{Digital Lab} \\
\textit{Vodafone}\\
Madrid, Spain \\
\href{https://orcid.org/0000-0002-9724-0099}{ORCID}}
\and
\IEEEauthorblockN{Carlos Carrasco}
\IEEEauthorblockA{\textit{Digital Lab} \\
\textit{Vodafone}\\
Madrid, Spain \\
\href{https://orcid.org/0000-0002-3569-6920}{ORCID}
}
}

\maketitle

\begin{abstract}
Nowadays, the way in which the people interact with computers has changed. Text- or voice-based interfaces are being widely applied in different industries. Among the most used ways of processing the user input are those based on intents or retrieval algorithms. In these solutions, important information of the user could be lost in the process. 
For the proposed natural language processing pipeline the entities are going to take a principal role, under the assumption that entities are where the purpose of the user resides. Entities fed with context will be projected to a specific domain supported by a knowledge graph, resulting in what has been named as linked entities. These linked entities serve then as a key for searching a top level aggregation concept within our knowledge graph.
\end{abstract}


\section{Introduction}

The problem of interacting with human users in order to process natural language, extract valuable information and providing sensible information in return, is not trivial.

During the last years, huge improvement has been achieved in terms of Human Computer Interfaces (HCI), in particular within the field of Natural Language Processing (NLP). The increasing computing resources, availability of data and connectivity quality, have empowered the need for automation of sales, technical support processes and health assistance among others.


As individuals express themselves in different ways, the ability human beings have for understanding each others is a difficult feature to be transferred to machines. A key task for linguistics resides in the fact that the meaning of words heavily depends on the context. This problem has several peculiarities which need to be treated in depth.

ISA-bEL, a dedicated NLP pipeline, Fig. \ref{fig:highlevelp}, is proposed with the objective of extracting meaning from different entities. This meaning will be given by the domain and the context of the sentence, resulting in entity linking. For example, the use case of processing a customer input, looking for the most suitable product or set of products, and wrapping them as an existing top level category called package.


In terms of the development framework, open source technologies has been chosen to enable deep customization. In this sense Rasa Open Source has been chosen as main platform, together with some broadly used machine learning and NLP libraries as SpacyNLP and Scikit-Learn. 

In this paper, a whole NLP pipeline will be defined and reviewed, from simple input processing stages to in-depth Natural Language Understanding (NLU) blocks which compose this engine. Moreover, the algorithms involved in later stages of the pipeline, databases and overall architecture will be reviewed for discussion.

\section{Background}

NLP is currently one of the most covered topics among technology forums, discussions and research conducted by artificial intelligence engineers.\cite{usagesnlpdeeplearning} However, the definition of NLP and its differentiation with other related topics as NLU remains unclear.


NLP can be defined as a subset of Artificial Intelligence (AI) which \textbf{processes} natural language (speech, text, etc.) and turns it into structured data. Nevertheless, NLU involves the \textbf{understanding} on what was meant by the input user. Similarly, Natural Language Generation (NLG), focuses on artificial \textbf{generation} of text, usually as a related response to a user input or given context. \cite{nlp_stateart}
\cite{nlpareview}

From now on in this paper, NLU and NLG will be interpreted as a subset of NLP, which is at the same time a subset of AI.

Current NLP solutions take advantage of state of the art computing techniques and algorithms, especially machine learning and deep learning algorithms \cite{10.5555/3110856}. In the last years, NLP techniques have used Recurrent Neural Networks (RNN)\cite{tarwani2017survey}, Long Short Term Memory Networks (LSTM)\cite{DBLP:journals/corr/abs-1909-09586}, Gated Recurrent Units (GRU), and more recently attention and transformers.\cite{NIPS2017_3f5ee243}\cite{Galassi20214291}


The processing pipeline is formed by every NLP/NLU operation necessary to understand the user's intent.

The typical first operation is tokenization, that means dividing up the input text into subunits, called tokens. This segmentation into tokens relies on a kind of pattern recognition involving collocational patterns \cite{webster1992tokenization}. So, a token represents the individual appearance of a word, number or punctuation mark in a certain position in a text \cite{Grefenstette1999}.

The part-of-speech tag (noun, verb, conjunction  and so on) of a given word in a given sentence is extracted by the part of speech tagging. 

The dependency parser is a syntactic analysis used to discover syntactic relations between individual tokens in a sentence and connects syntactically term-term relationships. \cite{vasiliev2020natural}


Information Extraction (IE) tasks are focused on extracting structured information from unstructured text.
Named Entity Recognition (NER) is one of the important sub-tasks of IE which recognizes entities. They may be persons, organizations, location names, times or dates in text. \cite{nadeau2007survey}


\section{Related work}

\begin{figure*}[!ht]
  \includegraphics[width=\textwidth]{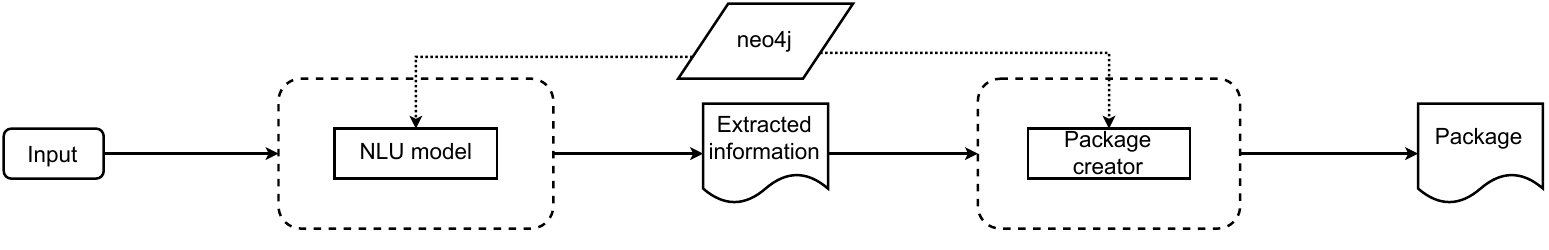}
  \caption{High level ISA-bEL pipeline.}
  \label{fig:highlevelp}
\end{figure*}

\begin{table*}

\caption{Examples of user interaction.}
\label{table:examples}
  \centering   
  \begin{tabularx}{\textwidth}[c]{XX}
    
    \textbf{\textcolor{blue}{Input}} & \textbf{\textcolor{blue}{Package}} \\ 
    \hline
    \noalign{\smallskip}
    \begin{minipage}[t]{\linewidth}%
    1) I want a computer with 1 TB of storage, graphic card to play videogames and shoes. 
    \end{minipage}
     & 
    \begin{minipage}[t]{\linewidth}%
    \textbf{Gaming advanced}: \\
    \{RAM\_16GB; CPU\_I7; GRAPHIC\_3080; STORAGE\_1TB\}
    \end{minipage}
    \\
    \noalign{\smallskip}
    \noalign{\smallskip}
    \hline
    \noalign{\smallskip}
    \begin{minipage}[t]{\linewidth}%
    2) I want a computer with 512 GB of storage and 8GB of RAM memory 
    \end{minipage}&
    \begin{minipage}[t]{\linewidth}%
   
    \textbf{Gaming beginner}: \\
    \{RAM\_8GB; CPU\_I3; GRAPHIC\_3060; STORAGE\_512GB\} \\
    \\
    \textbf{Gaming medium}: \\
    \{RAM\_8GB; CPU\_I5; GRAPHIC\_3070; STORAGE\_512GB\}
    
    \end{minipage}   
    \\
    \noalign{\smallskip}
    \noalign{\smallskip}
    \hline
    \noalign{\smallskip}
    3) I want a computer with i5 processor, 512 GB of storage and 8GB of RAM memory 
    &  
    \begin{minipage}[t]{\linewidth}%
    \textbf{Gaming medium}:\\
    \{RAM\_8GB; CPU\_I5; GRAPHIC\_3070; STORAGE\_512GB\}
    
    \end{minipage} 
    \\
    
    \noalign{\smallskip}
    \hline
    \noalign{\smallskip}  
    \\

\end{tabularx}

\end{table*}

 Before diving deeper on our proposed method, it is important to do some research on similar findings and developments, about the interaction with users through natural language processing, and the ways in which these systems return relevant info to the users.
 
The solutions which use natural language are growing in the last decades. The ripeness of this technology is turning into solid application. The possibilities of this technology could be:  sentiment analysis, text classification, search engines, chatbots, assistants, summarizing, translation, predictive text, correction, natural language generation, etc \cite{yadav2020sentiment} \cite{9076358}.

This technology offers a huge range of use cases being at the center of our attention those that interact  with persons directly (maintaining a conversation like a chatbot) or indirectly (analyzing users texts).

The solutions to be reviewed can be grouped in three main clusters \cite{overview_chatbot_2020}:
\begin{itemize}
    \item Rule-based: They establish a bidirectional conversation mostly based on hand-code rules.
    \item Retrieval-based: Their objective is to get back the information request.
    \item Generative: Under an user constrains the algorithm generate an output based on its knowledge.
\end{itemize}

\subsection{Rule-based}
From the point of view of NLP developments as human-computer interfaces, one of the recently and widely adopted solutions are NLP/NLU chatbots \cite{survey_design_2017}. These (usually) include several distinguished blocks or modules \cite{overview_chatbot_2020}:
\begin{itemize}
    \item NLP stages
    \item NLU engine
    \item Chatbot engine (response selector, interactions, etc.)
\end{itemize}
These modules should work together for the chatbot to provide highly human-like interactions with the users. The internal NLU engine will provide better results if the input processing is done in a proper way for the problem which the system is handling. Thus, the response provided by the chatbot will be more sensible to the user. 

Normally, regular expressions, slots, spreadsheets and similar tools are used for information extraction, having the following counterpart: the sentence structure should always be similar for them to work properly.

Most cloud vendors as Amazon Web Services (AWS), Google Cloud Platform (GCP), Microsoft Azure or IBM, include these conversational services in their portfolios. This cloud-based frameworks are closed platform \cite{overview_chatbot_2020} and offer from simple NLU engines to complete assistants as AWS Comprehend\footnote{\url{https://aws.amazon.com/es/comprehend}}, GCP Dialogflow\footnote{\url{https://dialogflow.com}}, Azure Bot Service or IBM Watson \footnote{\url{https://www.ibm.com/watson}}, each one exploiting their own NLP engine which is usually limited in terms of customization \cite{MALAMAS2021109} and adaptation to more specific tasks, but powerful in terms of knowledge for general purposes and its development process is easy and quick. 


However, there exist open-source platforms with full customization possibilities and adaptation to distinct problems, allowing engineers to intervene in most aspects of development. 

\subsection{Retrieval-based}

Question answering (QA) is an important task in NLP, these algorithms aim to provide precise answers based on large-scale unstructured documents. To achieve it, the algorithm has to deal with two critical points: understanding the user and finding the response in the documents \cite{zhu2021retrieving}.

This algorithms are changing concurrently with NLP techniques improvements, being a long journey which starts with question answering in baseball \cite{green1961baseball}, to the most recent and complex techniques using GPT-3 to address this problem \cite{nakano2021webgpt}

There exist two branches within QA: close-domain and open-domain. The close-domain system deal with questions under a specific domain, it can exploit domain specific knowledge by using a model fitted to a unique-domain database \cite{woods1997conceptual} \cite{molla2007question} \cite{green1961baseball}. On the other side, open-domain systems deal with questions about nearly anything and can only rely on knowledge graphs \cite{harabagiu2003open} \cite{Chen20171870}. 

Knowledge graphs are used to store entities, relationships between entities and their properties \cite{Ehrlinger2016}. The entities  could be for a specific domain or for organizations. The most common knowledge graphs are YAGO \cite{Suchanek2007697}, DBPedia \cite{Auer2007722} or Wikidata \cite{Vrande201478}.

The open-domain QA is normally a more complex system where solutions are usually tuned to meet specific needs (i.e. a QA algorithm that processes the question twice looking for better results) \cite{qi2019answering}, but it is known that open and close domain should cover specific tasks to answer the user.
\begin{itemize}
    \item Retriever: This part is aimed to obtained a number of relevant document that probably contain the correct answer.
    \cite{yang2019end} \cite{nishida2018retrieve} 
    \item Reader: It has the intention to return the answer to  the question.  \cite{rajpurkar2016squad} \cite{hermann2015teaching}
\end{itemize}

Recently, QA is changing because of the improvements on transformer-based algorithms, giving the possibility of grouping different tasks to be solved at the same time \cite{zhu2021retrieving}: retriever-reader \cite{wang2018r} , retriever-only \cite{seo2019real} and retriever-free  \cite{lewis2019bart}.

\subsection{Generative}
\label{sec:generation}


In most cases, generative models are based on Transformer architecture. Their most common NLP objective is solving sequence-to-sequence tasks \cite{raffel2019exploring}. Some of these tasks are: question answering, document summarizing, sentiment classification or translation \cite{raffel2019exploring}.


The use of these model has recently grown thanks to: pretrainned models and facilities to build high-capacity models \cite{wolf2020transformers}. 

The solutions for different use cases generally need a huge amount of data and a big computation cost \cite{raffel2019exploring}. For this reason, the transfer learning is helping its use, reducing the cost of train this algorithms \cite{pan2009survey}.

Similar to the solution that we purpose is the creation of cooking recipe from a list of ingredient: RecipeNLG \cite{bien2020recipenlg} and Chef Transformer \cite{cheftransformer}. Theses algorithms use more than 2 millions of recipe to produce a good performance. 

For the correct use of artificial intelligence models in production environment, the focus must also be placed on other objective beyond the precision. Explainable artificial intelligence is a crucial feature for the practical deployment of AI models \cite{arrieta2020explainable} being the most important terms: understandability, comprehensibility, interpretability, explainability and transparency.

The deep neural network use to solve the majority of the problems cause of the effectiveness. They are been considered "black-boxes" for no cover correctly the previous topics. Model-agnostic are growing to help in the explainability and understandability of these model such as: LIME \cite{lime} and SHAP \cite{NIPS2017_7062}. The visualizations designed to explore the Transformer architectures enable some additional features like: exploration of all neuronal cells or attention maps \cite{9373074}.

\begin{figure}[h]
    \includegraphics[width=\linewidth]{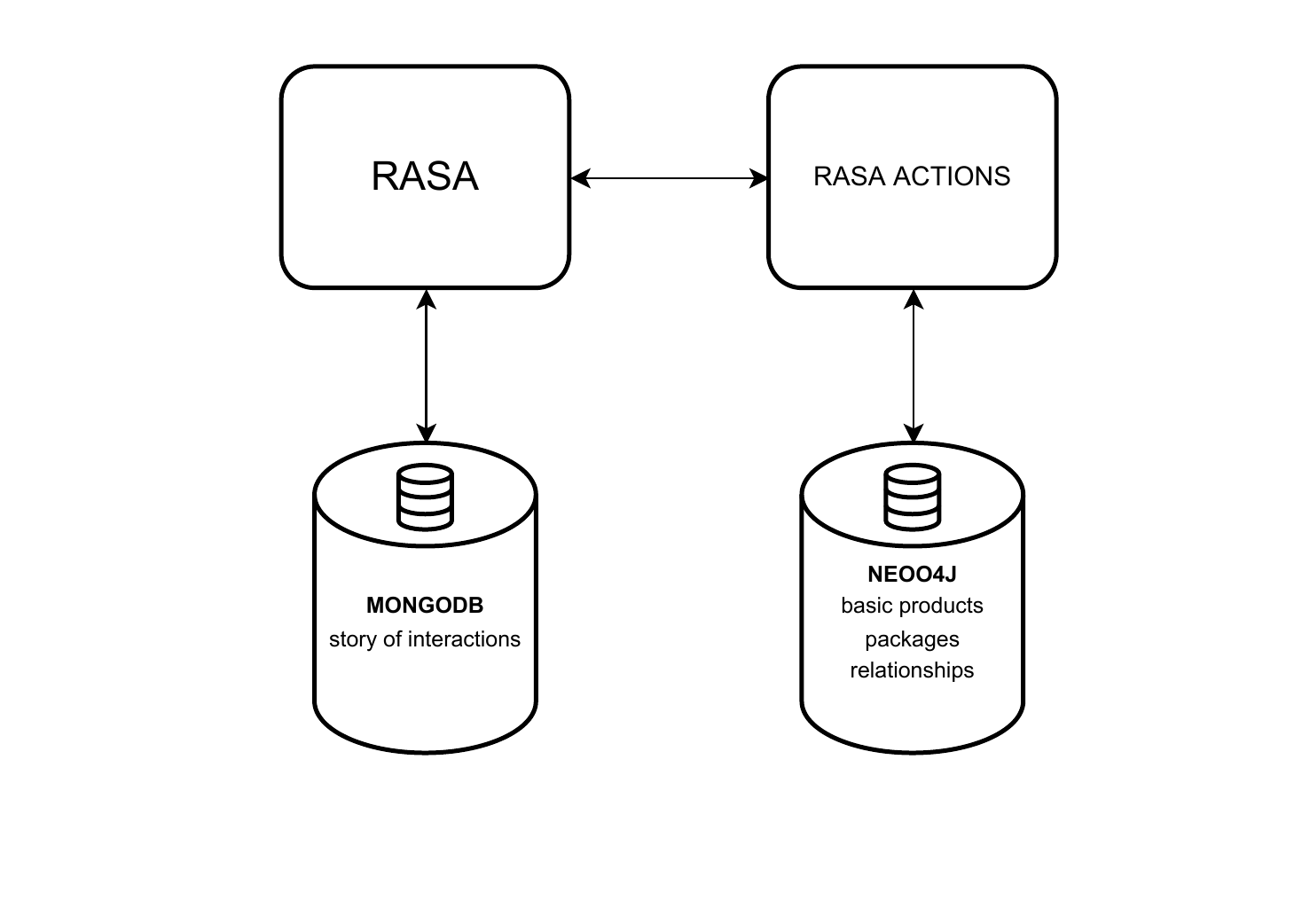}
    \caption{ISA-bEL. High level architecture diagram.}
    \label{fig:highlevelarch}
\end{figure}

\section{Methodology}
\label{sec:methodology}
\begin{figure*}[ht]
  \includegraphics[width=\textwidth]{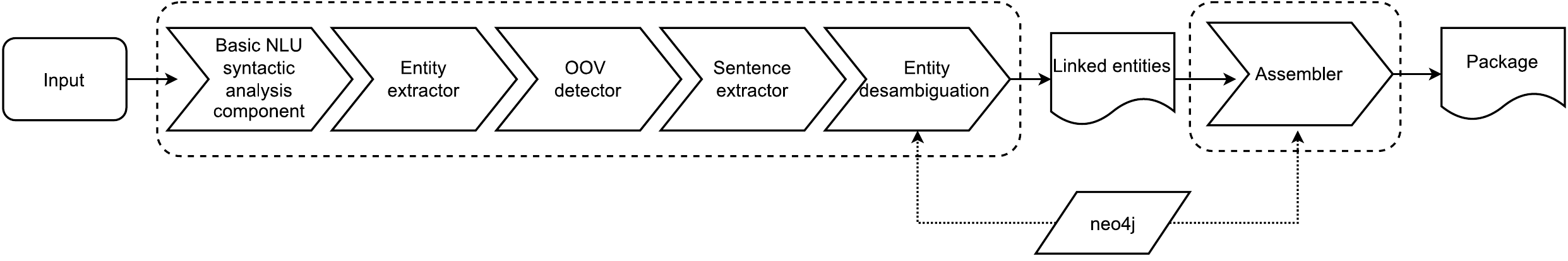}
  \caption{Macro ISA-bEL pipeline.}
  \label{fig:isapipe}
\end{figure*}

The scope of the research was to develop a tool which would receive a user input, detect N entities included in the request, process them and generate a package which would suit best the user needs, Fig. \ref{fig:highlevelp}. Of course, it would comply with the restrictions which implicitly lie on the \textbf{relationships} between different linked entities and packages.

This tool has been developed as a cloud service, exposed to the internet as an API, wrapped as a complete solution. For this purpose, a set of containers has been deployed to a cloud platform, each one including a basic module of the system.

Rasa \cite{bocklisch2017rasa} was chosen as main framework, mainly for being an open source environment, fully customizable whether at the NLU engine, preprocessing stages or chatbot interaction blocks. Moreover, a custom spanish SpacyNLP\footnote{\url{https://spacy.io/}} model has been trained and embedded into a Rasa pipeline, exploiting both SpacyNLP and Rasa capabilities together.

With Rasa (Open Source version) almost every detail can be tailored to specific needs. First of all, Rasa offers built-in components for the NLP pipeline. These include tokenizers, grammar correctors, featurizers, entity extractors and other chatbot-oriented modules as classifiers, response selectors, rules, etc. Moreover, the underlying NLP model can be trained either from rasa CLI or as a MitieNLP \cite{dlib09} or SpacyNLP model in a separate python pipeline, which allows the developer to reach the lowest levels of the NLU blocks and features. Thus, Rasa offers the possibility to design a complete pipeline which includes the processing stages (NLP), the NLU blocks which conform the NLU engine as a trainable model, and the interaction blocks if an assistant or chatbot functionality is desired. In addition to the pipeline, Rasa offers an action server which is able to execute custom code blocks whenever a specific functionality is needed after being triggered from the NLP pipeline.

In our case, a pre-trained SpacyNLP model and the different components it includes, have been trained in a separate python pipeline and embedded into a Rasa pipeline as the underlying language model. With this, the highest customization is achieved, so that the system is specialized in this specific task. The most important SpacyNLP component which has been trained is the NER, and so it will be revisited in section \hyperref[sec:isabel]{ISA-bEL}. Components as the Dependency Parser could be trained to improve the performance. However, this escapes the goal of this research.

As a high level architecture overview, four containers would be necessary, as depicted in Fig. \ref{fig:highlevelarch}. Two of them conform the Rasa part of the solution, one for the Rasa server and one for the actions server respectively. The other two containers are databases. The first one for a MongoDB database which stores the story of conversations and interactions with user. The last container is a graph database (Neo4J) which stores linked entities, packages and relationships between them. This fourth container acts as the knowledge graph of the solution.

\section{ISA-bEL}
\label{sec:isabel}

\begin{figure}[h!]
  \includegraphics[width=0.5\textwidth]{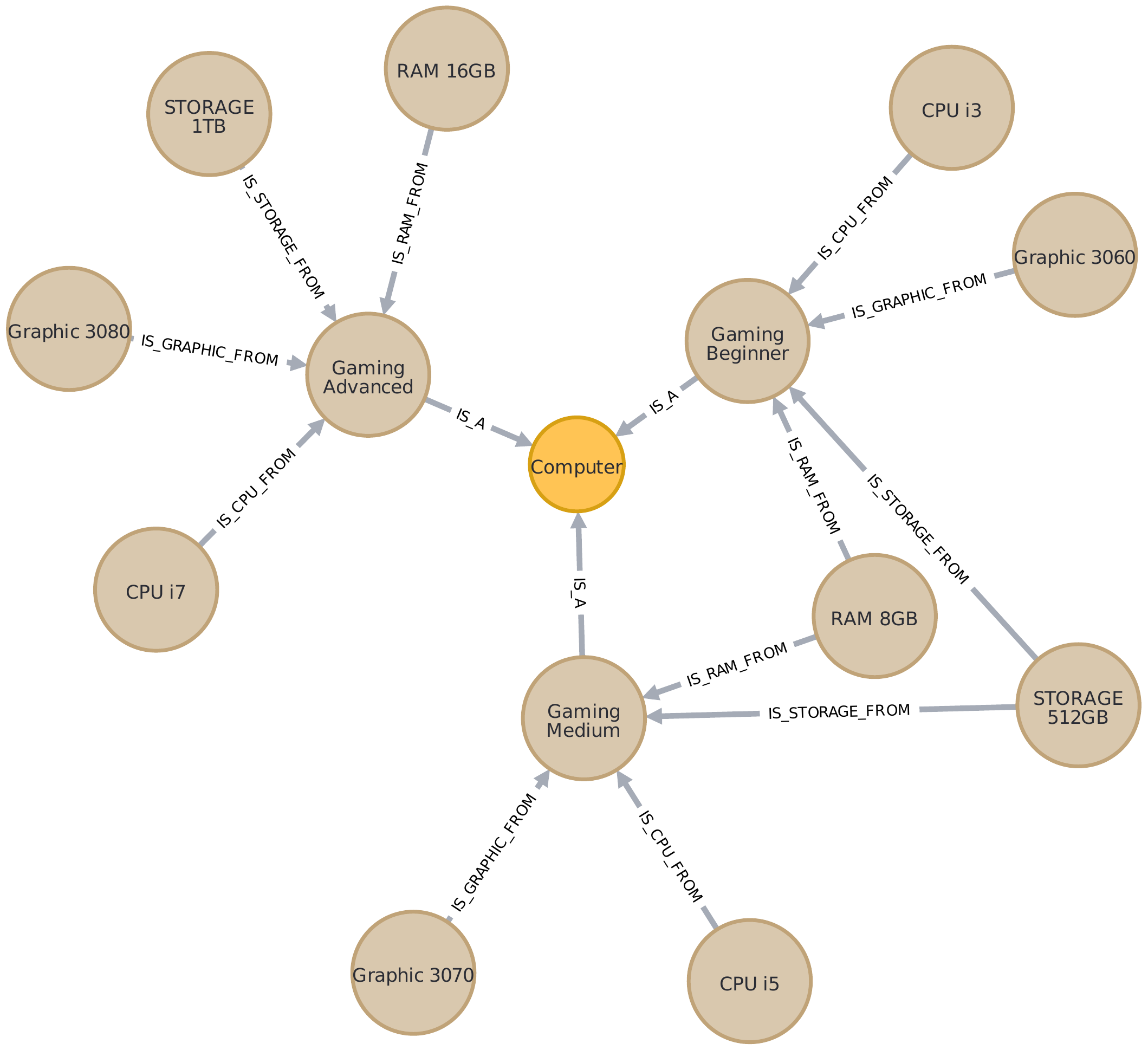}
  \caption{Knowledge Base Graph.}
  \label{fig:kgb}
\end{figure}

To understand the proposed pipeline, depicted in  Fig. \ref{fig:isapipe}, the explanation will be supported by a computer configuration example, in which the contribution of each component of the pipeline to the example case will be reviewed. The example to be used in the following subsections is the first one appearing at Table \ref{table:examples}. The Knowledge Base Graph in Fig. \ref{fig:kgb} is in which the solutions given for the examples in this paper are based.


\subsection{Entity extractor}
\label{sec:detectentities}  
Our proposed methodology has a strong dependency on named entity recognition. Considering that entities help to determine the user intention, this processing stage is one of the most critical points of ISA-bEL.
The detection of entities could be performed by more than one module. These modules will be executed sequentially. The first module that classifies a word into an entity will have the priority. Two paths could be followed for entity extraction:
\begin{itemize}
    \item Pattern-based: These modules are completely based on pattern-recognition and pattern-matching \cite{survey_design_2017}, which are indeed hand-code methods for this task. However, the effectiveness it provides for specific words whose meaning does not depend on the context will be higher.
    \item NER model: These methods detect an entity through the execution of an artificial intelligence algorithm \cite{li2020survey}. There exist different pre-trained models for different intention. Furthermore, a more specialized one could be trained for a specific task.
\end{itemize}

While pattern-based methods are more accurate in terms of the definition of entities and words they detect, they could lead to problems with words with different meanings because a single meaning of a word is detected. 
On the other side, when NER algorithms are used to detect entities depending on the context like a city in the sentence ``I want to travel to...". This could lead to words detected as an entity which in fact do not belong to the domain.

\begin{example}
\label{ex-input}
The existing entities at the user input would be extracted and returned as the output of this component. They have been highlighted in gray at the input of our example in Table \ref{table:examples}: \\
\[
\text{I want a computer with } \text{1 TB} \overbrace{\text{\colorbox{gray!30}{storage}}}^{\text{entity 1}},
\]
\[
\overbrace{\text{\colorbox{gray!30}{graphic card}}}^{\text{entity 2}} \text{to play videogames} \text{ and} \overbrace{\text{\colorbox{gray!30}{shoes}.}}^{\text{entity 3}}
\]

\end{example}

\subsection{Out of vocabulary (OOV) detector}
To minimize the impact of detecting words which do not belong to the domain as entities, this module checks if every entity is included in a dictionary which includes every word that belongs to the domain, also called bag of words.

Performing this kind of scanning is not trivial, therefore some corrections should be applied. Thus, the proposed task would be the following \cite{bocklisch2017rasa}:
\begin{enumerate}
    \item Transforming to a comparable codification like deleting accents, or transforming to lowercase. Moreover, it is not possible to define every number, so numbers would be encoded as a constant value. 
    \item Extracting the lemma of the word simplifying word storage and covering more vocabulary.
    \item Finally, checking if the word exists.
\end{enumerate}

This step is simple in terms of complexity, but it is an important one for reducing the errors in the outcome of the model and making it more resilient. It gives ISA-bEL the power to better know its knowledge boundaries and show the user where these limits are.

\begin{example}
\label{ex-oov}
The context of computer does not include shoes. For this reason, the entity \textit{shoes} should not be taken into account. Therefore, the output of this component would ommit this entity, which was detected at Example \ref{ex-input}. \\
\[
\text{I want a computer with } \text{1 TB} \overbrace{\text{\colorbox{gray!30}{storage}}}^{\text{entity 1}},
\]
\[
\overbrace{\text{\colorbox{gray!30}{graphic card}}}^{\text{entity 2}} \text{to play videogames} \text{ and} \overbrace{\text{\colorbox{gray!30}{shoes}.}}^{\text{OOV}}
\]

\end{example}

\subsection{Sub-sentence extractor}
When an entity is detected, the next step is to disambiguate its meaning in order to establish a link between them and existing entities stored in the knowledge base graph. 

Sub-sentences would be created around the entities, adding other words of the input which have important syntactic or semantic relation with the entity being processed. This implies enriching the step of disambiguation and improving its performance.

Specific entities are chosen to create this sub-sentence because not all entities have the same importance. In general, the chosen entities would be those present in the knowledge graph.

The important components that would be needed for this creation are the dependency parser \cite{chen2014fast} and chunks \cite{abney1991parsing}. These together could help establishing relationships between entities and the rest of the words around them, so the similarity algorithm used in the next stage can take the context into account.

\begin{example}
\label{ex-subsentences}
The ``storage'' entity is incremented by the words ``1 TB'' and the entity ``graphic card'' is incremented by the words ``to play videogames'', in both cases these surrounding words complement the entities extracted at Example \ref{ex-oov} after removing OOV words.\\

\[
\text{I want a computer with } \underbracket{\text{1 TB} \overbrace{\text{\colorbox{gray!30}{storage}}}^{\text{entity 1}}}_{\text{sub-sentence A}},
\]
\[
\underbracket{\overbrace{\text{\colorbox{gray!30}{graphic card}}}^{\text{entity 2}} \text{to play videogames}}_{\text{sub-sentence B}} \text{and} \overbrace{\text{\colorbox{gray!30}{shoes}.}}^{\text{OOV}}
\]

\end{example}

\subsection{Entity Disambiguation}

The previous sub-sentences will be used to disambiguate the entities using a similarity algorithm to obtain linked entities from a chosen KG \cite{zhu2018exploiting}. At this stage, it is important to know that each sub-sentence could return more than one candidate.

A way to perform this is computing the cosine similarity between two sentences using their encoding \cite{reimers2019sentence}. After having generated the sub-sentences as described in previous sections, the best candidates for each sub-sentence is determined from the existing KG information.

This task could be done by the "EntityLinker" component in Spacy, this component is prepared to disambiguate an entity using the entire sentence with the SpaCy knowledge base that already has the entities linked to possible definitions.

\begin{example}
\label{ex-links}
The sub-sentences generated in Example \ref{ex-subsentences} will be compared with the entity description in database applying cosine similarity .
In this case, entitites "storage 1tb" and "graphic card 3080" are matched.\\

\[
\text{I want a computer with } \overbracket{\text{1 TB} \colorbox{gray!30}{storage}}^{\text{\textless KG:STORAGE\_1TB \textgreater}},
\]
\[
\overbracket{\text{\colorbox{gray!30}{graphic card}} \text{ to play videogames}}^{\text{\text{\textless KG:GRAPHIC\_3080 \textgreater}}} \text{and} \overbrace{\text{\colorbox{gray!30}{shoes}}.}^{\text{OOV}}
\]

\end{example}

\subsection{Assembler}

To fulfill the objective of this task, only linked entities and the KG are used to search and assemble the outcome for the user. 

Our approach in this point is to find the higher-level category which can group the linked entities. Returning this top category and the linked entities beneath it in the hierarchy.

The task of defining use cases could be considered an extra. The only target here has been to link entities. On the same trend, this task could be more necessary and complex if adding more categories and rules in the relationship. 





\begin{example}
Thanks to the entity link detected in \ref{ex-links}, and after going one level higher, the graph determines that the user wants an advanced gaming computer, together with a "CPU i7" and a "RAM 16 GB".\\
    \begin{figure}[h]
      \includegraphics[width=0.5\textwidth]{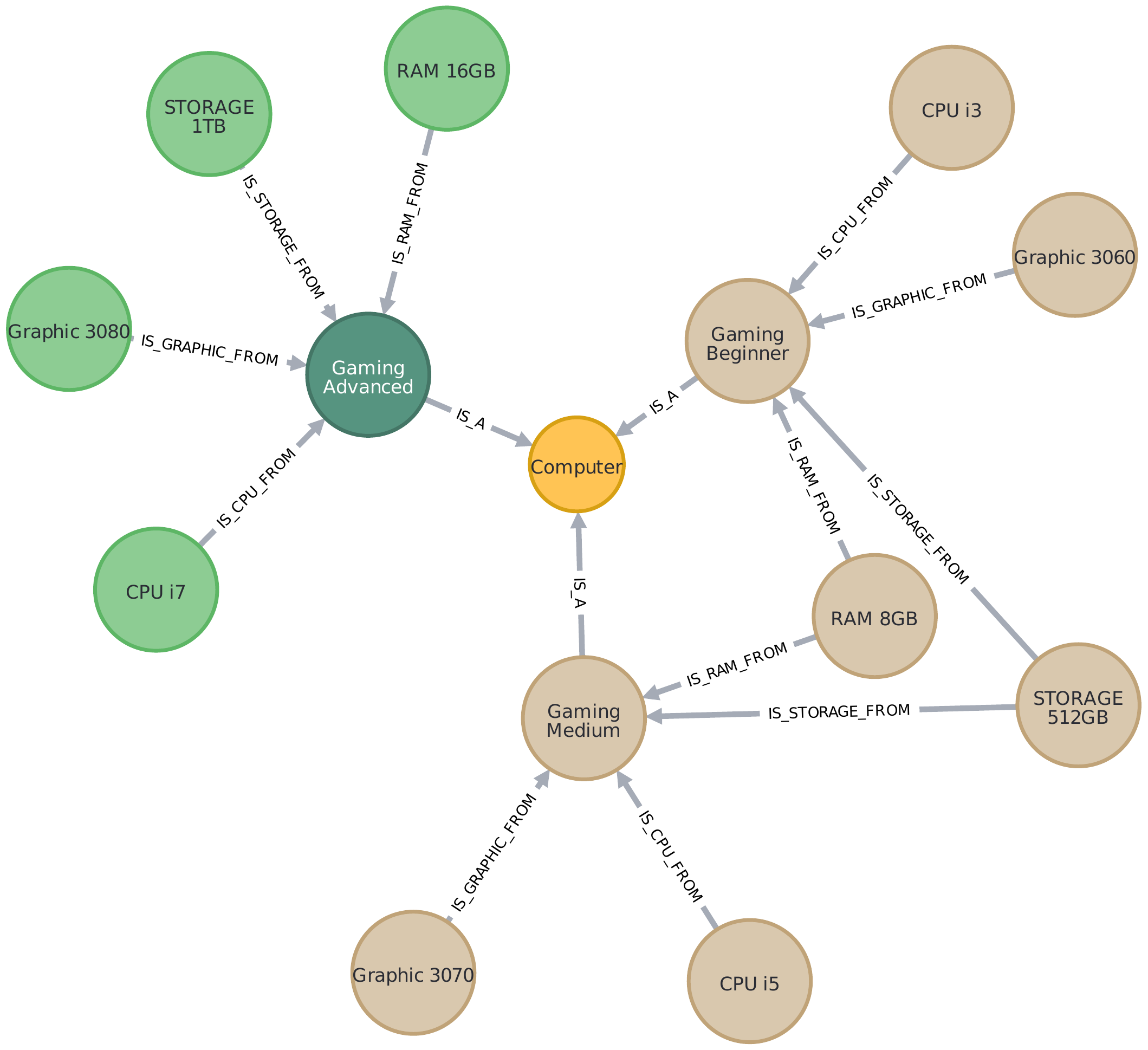}
      \caption{Aggregation in top category}
    \end{figure}
\end{example}

\section{conclusions and future work}

First of all, one of the most important limitations is the language in which this research has been developed, Spanish. Despite being one of the most spoken languages in the world, there are fewer libraries and developments which use Spanish as their base languages. For example, English has a stronger community support. Thus, some building blocks of our pipeline could improve their performance if trained in English. However, some pre-trained components in Spanish have been found, re-trained and/or used.

In addition to language limitations, this research has not included the contribution of any linguistics. This means that certain grammatical tasks of the pipeline could be improved with the help of specialists in this field.

Another limitation is the fact that Rasa detects entities and intents via an entity extractor (i.e. DIETClassifier). This is the component used for triggering default actions as responding, waiting, asking or custom actions as querying a database or running some code block. Thus, the execution of custom actions is limited to the intents which Rasa detects, triggered by rasa-labeled intents inside Rasa domain. Spacy has no influence here, and this fact is a stopper. 

Despite these limitations, the pipeline has proved to perform properly in terms of entity extraction and linking. The solution has been developed in a closed domain task for entity disambiguation simplicity.

Due to the open source nature of this solution, it is a fully scalable system, both from the framework in which it is wrapped and from the customization point of view. The first four components of the pipeline are shown in Fig. \ref{fig:isapipe}: entity extractor, OOV detector, sentence extractor and entity disambiguation have been designed to solve the problem of understanding the user, the latter being the one which allows adaptation to a specific use case.

Comparing it with the recipe use cases mentioned at \hyperref[sec:generation]{generative section at related work}, ISA-bEL does not create or generate something new, it is always based on the knowledge graph content. For this reason, it is critical to build a robust, well designed KG.

To sum up previous aspects, a dynamic system which adapts to a user input without applying fix rules as conventional chatbot engines has been presented, developed and reviewed. With this work, the gap between academic research and production systems has been narrowed, remaining at the same time open for further research and implementations.

\bibliographystyle{ieeetr} 
\bibliography{citas} 

\begin{thebibliography}{10}

\bibitem{usagesnlpdeeplearning}
D.~W. Otter, J.~R. Medina, and J.~K. Kalita, ``A survey of the usages of deep
  learning in natural language processing,'' {\em CoRR}, vol.~abs/1807.10854,
  2018.

\bibitem{nlp_stateart}
D.~Khurana, A.~Koli, K.~Khatter, and S.~Singh, ``Natural language processing:
  State of the art, current trends and challenges,'' 08 2017.

\bibitem{nlpareview}
S.~Joseph, K.~Sedimo, F.~Kaniwa, H.~Hlomani, and K.~Letsholo, ``Natural
  language processing: A review,'' {\em Natural Language Processing: A Review},
  vol.~6, pp.~207--210, 03 2016.

\bibitem{10.5555/3110856}
Y.~Goldberg and G.~Hirst, {\em Neural Network Methods in Natural Language
  Processing}.
\newblock Morgan \& Claypool Publishers, 2017.

\bibitem{tarwani2017survey}
K.~M. Tarwani and S.~Edem, ``Survey on recurrent neural network in natural
  language processing,'' {\em Int. J. Eng. Trends Technol}, vol.~48,
  pp.~301--304, 2017.

\bibitem{DBLP:journals/corr/abs-1909-09586}
R.~C. Staudemeyer and E.~R. Morris, ``Understanding {LSTM} - a tutorial into
  long short-term memory recurrent neural networks,'' {\em CoRR},
  vol.~abs/1909.09586, 2019.

\bibitem{NIPS2017_3f5ee243}
A.~Vaswani, N.~Shazeer, N.~Parmar, J.~Uszkoreit, L.~Jones, A.~N. Gomez, L.~u.
  Kaiser, and I.~Polosukhin, ``Attention is all you need,'' vol.~30, 2017.

\bibitem{Galassi20214291}
A.~Galassi, M.~Lippi, and P.~Torroni, ``Attention in natural language
  processing,'' {\em IEEE Transactions on Neural Networks and Learning
  Systems}, vol.~32, no.~10, pp.~4291--4308, 2021.
\newblock cited By 25.

\bibitem{webster1992tokenization}
J.~J. Webster and C.~Kit, ``Tokenization as the initial phase in nlp,'' in {\em
  COLING 1992 Volume 4: The 14th International Conference on Computational
  Linguistics}, 1992.

\bibitem{Grefenstette1999}
G.~Grefenstette and H.~van Halteren, {\em Tokenization}, pp.~117--133.
\newblock Dordrecht: Springer Netherlands, 1999.

\bibitem{vasiliev2020natural}
Y.~Vasiliev, {\em Natural Language Processing with Python and SpaCy: A
  Practical Introduction}.
\newblock No Starch Press, 2020.

\bibitem{nadeau2007survey}
D.~Nadeau and S.~Sekine, ``A survey of named entity recognition and
  classification,'' {\em Lingvisticae Investigationes}, vol.~30, no.~1,
  pp.~3--26, 2007.

\bibitem{yadav2020sentiment}
A.~Yadav and D.~K. Vishwakarma, ``Sentiment analysis using deep learning
  architectures: a review,'' {\em Artificial Intelligence Review}, vol.~53,
  no.~6, pp.~4335--4385, 2020.

\bibitem{9076358}
Rahul, S.~Adhikari, and Monika, ``Nlp based machine learning approaches for
  text summarization,'' in {\em 2020 Fourth International Conference on
  Computing Methodologies and Communication (ICCMC)}, pp.~535--538, 2020.

\bibitem{overview_chatbot_2020}
E.~Adamopoulou and L.~Moussiades, ``An overview of chatbot technology,'' in
  {\em Artificial Intelligence Applications and Innovations} (I.~Maglogiannis,
  L.~Iliadis, and E.~Pimenidis, eds.), (Cham), pp.~373--383, Springer
  International Publishing, 2020.

\bibitem{survey_design_2017}
K.~Ramesh, S.~Ravishankaran, A.~Joshi, and K.~Chandrasekaran, ``A survey of
  design techniques for conversational agents,'' in {\em Information,
  Communication and Computing Technology} (S.~Kaushik, D.~Gupta, L.~Kharb, and
  D.~Chahal, eds.), (Singapore), pp.~336--350, Springer Singapore, 2017.

\bibitem{MALAMAS2021109}
N.~Malamas and A.~Symeonidis, ``Embedding rasa in edge devices: Capabilities
  and limitations,'' {\em Procedia Computer Science}, vol.~192, pp.~109--118,
  2021.
\newblock Knowledge-Based and Intelligent Information \& Engineering Systems:
  Proceedings of the 25th International Conference KES2021.

\bibitem{zhu2021retrieving}
F.~Zhu, W.~Lei, C.~Wang, J.~Zheng, S.~Poria, and T.-S. Chua, ``Retrieving and
  reading: A comprehensive survey on open-domain question answering,'' {\em
  arXiv preprint arXiv:2101.00774}, 2021.

\bibitem{green1961baseball}
B.~F. Green~Jr, A.~K. Wolf, C.~Chomsky, and K.~Laughery, ``Baseball: an
  automatic question-answerer,'' in {\em Papers presented at the May 9-11,
  1961, western joint IRE-AIEE-ACM computer conference}, pp.~219--224, 1961.

\bibitem{nakano2021webgpt}
R.~Nakano, J.~Hilton, S.~Balaji, J.~Wu, L.~Ouyang, C.~Kim, C.~Hesse, S.~Jain,
  V.~Kosaraju, W.~Saunders, {\em et~al.}, ``Webgpt: Browser-assisted
  question-answering with human feedback,'' {\em arXiv preprint
  arXiv:2112.09332}, 2021.

\bibitem{woods1997conceptual}
W.~A. Woods, {\em Conceptual indexing: A better way to organize knowledge}.
\newblock Sun Microsystems, Inc., 1997.

\bibitem{molla2007question}
D.~Moll{\'a} and J.~L. Vicedo, ``Question answering in restricted domains: An
  overview,'' {\em Computational Linguistics}, vol.~33, no.~1, pp.~41--61,
  2007.

\bibitem{harabagiu2003open}
S.~M. Harabagiu, S.~J. Maiorano, and M.~A. Pa{\c{s}}ca, ``Open-domain textual
  question answering techniques,'' {\em Natural Language Engineering}, vol.~9,
  no.~3, pp.~231--267, 2003.

\bibitem{Chen20171870}
D.~Chen, A.~Fisch, J.~Weston, and A.~Bordes, ``Reading wikipedia to answer
  open-domain questions,'' in {\em Reading Wikipedia to answer open-domain
  questions}, vol.~1, pp.~1870--1879, 2017.
\newblock cited By 424.

\bibitem{Ehrlinger2016}
L.~Ehrlinger and W.~Wöß, ``Towards a definition of knowledge graphs,'' in
  {\em Towards a definition of knowledge graphs}, vol.~1695, 2016.
\newblock cited By 19.

\bibitem{Suchanek2007697}
F.~Suchanek, G.~Kasneci, and G.~Weikum, ``Yago: A core of semantic knowledge,''
  in {\em Towards a definition of knowledge graphs}, pp.~697--706, 2007.
\newblock cited By 2260.

\bibitem{Auer2007722}
S.~Auer, C.~Bizer, G.~Kobilarov, J.~Lehmann, R.~Cyganiak, and Z.~Ives,
  ``Dbpedia: A nucleus for a web of open data,'' {\em Lecture Notes in Computer
  Science (including subseries Lecture Notes in Artificial Intelligence and
  Lecture Notes in Bioinformatics)}, vol.~4825 LNCS, pp.~722--735, 2007.
\newblock cited By 2306.

\bibitem{Vrande201478}
D.~Vrandečić and M.~Krötzsch, ``Wikidata: A free collaborative
  knowledgebase,'' {\em Communications of the ACM}, vol.~57, no.~10,
  pp.~78--85, 2014.
\newblock cited By 1092.

\bibitem{qi2019answering}
P.~Qi, X.~Lin, L.~Mehr, Z.~Wang, and C.~D. Manning, ``Answering complex
  open-domain questions through iterative query generation,'' {\em arXiv
  preprint arXiv:1910.07000}, 2019.

\bibitem{yang2019end}
W.~Yang, Y.~Xie, A.~Lin, X.~Li, L.~Tan, K.~Xiong, M.~Li, and J.~Lin,
  ``End-to-end open-domain question answering with bertserini,'' {\em arXiv
  preprint arXiv:1902.01718}, 2019.

\bibitem{nishida2018retrieve}
K.~Nishida, I.~Saito, A.~Otsuka, H.~Asano, and J.~Tomita, ``Retrieve-and-read:
  Multi-task learning of information retrieval and reading comprehension,'' in
  {\em Proceedings of the 27th ACM international conference on information and
  knowledge management}, pp.~647--656, 2018.

\bibitem{rajpurkar2016squad}
P.~Rajpurkar, J.~Zhang, K.~Lopyrev, and P.~Liang, ``Squad: 100,000+ questions
  for machine comprehension of text,'' {\em arXiv preprint arXiv:1606.05250},
  2016.

\bibitem{hermann2015teaching}
K.~M. Hermann, T.~Kocisky, E.~Grefenstette, L.~Espeholt, W.~Kay, M.~Suleyman,
  and P.~Blunsom, ``Teaching machines to read and comprehend,'' {\em Advances
  in neural information processing systems}, vol.~28, 2015.

\bibitem{wang2018r}
S.~Wang, M.~Yu, X.~Guo, Z.~Wang, T.~Klinger, W.~Zhang, S.~Chang, G.~Tesauro,
  B.~Zhou, and J.~Jiang, ``R 3: Reinforced ranker-reader for open-domain
  question answering,'' in {\em Thirty-Second AAAI Conference on Artificial
  Intelligence}, 2018.

\bibitem{seo2019real}
M.~Seo, J.~Lee, T.~Kwiatkowski, A.~P. Parikh, A.~Farhadi, and H.~Hajishirzi,
  ``Real-time open-domain question answering with dense-sparse phrase index,''
  {\em arXiv preprint arXiv:1906.05807}, 2019.

\bibitem{lewis2019bart}
M.~Lewis, Y.~Liu, N.~Goyal, M.~Ghazvininejad, A.~Mohamed, O.~Levy, V.~Stoyanov,
  and L.~Zettlemoyer, ``Bart: Denoising sequence-to-sequence pre-training for
  natural language generation, translation, and comprehension,'' {\em arXiv
  preprint arXiv:1910.13461}, 2019.

\bibitem{raffel2019exploring}
C.~Raffel, N.~Shazeer, A.~Roberts, K.~Lee, S.~Narang, M.~Matena, Y.~Zhou,
  W.~Li, and P.~J. Liu, ``Exploring the limits of transfer learning with a
  unified text-to-text transformer,'' {\em arXiv preprint arXiv:1910.10683},
  2019.

\bibitem{wolf2020transformers}
T.~Wolf, L.~Debut, V.~Sanh, J.~Chaumond, C.~Delangue, A.~Moi, P.~Cistac,
  T.~Rault, R.~Louf, M.~Funtowicz, {\em et~al.}, ``Transformers:
  State-of-the-art natural language processing,'' in {\em Proceedings of the
  2020 conference on empirical methods in natural language processing: system
  demonstrations}, pp.~38--45, 2020.

\bibitem{pan2009survey}
S.~J. Pan and Q.~Yang, ``A survey on transfer learning,'' {\em IEEE
  Transactions on knowledge and data engineering}, vol.~22, no.~10,
  pp.~1345--1359, 2009.

\bibitem{bien2020recipenlg}
M.~Bie{\'n}, M.~Gilski, M.~Maciejewska, W.~Taisner, D.~Wisniewski, and
  A.~Lawrynowicz, ``Recipenlg: A cooking recipes dataset for semi-structured
  text generation,'' in {\em Proceedings of the 13th International Conference
  on Natural Language Generation}, pp.~22--28, 2020.

\bibitem{cheftransformer}
K.~G.~e. Mehrdad~Farahani, ``{Chef Transformer},'' 8 2021.

\bibitem{arrieta2020explainable}
A.~B. Arrieta, N.~D{\'\i}az-Rodr{\'\i}guez, J.~Del~Ser, A.~Bennetot, S.~Tabik,
  A.~Barbado, S.~Garc{\'\i}a, S.~Gil-L{\'o}pez, D.~Molina, R.~Benjamins, {\em
  et~al.}, ``Explainable artificial intelligence (xai): Concepts, taxonomies,
  opportunities and challenges toward responsible ai,'' {\em Information
  fusion}, vol.~58, pp.~82--115, 2020.

\bibitem{lime}
M.~T. Ribeiro, S.~Singh, and C.~Guestrin, ``"why should {I} trust you?":
  Explaining the predictions of any classifier,'' in {\em Proceedings of the
  22nd {ACM} {SIGKDD} International Conference on Knowledge Discovery and Data
  Mining, San Francisco, CA, USA, August 13-17, 2016}, pp.~1135--1144, 2016.

\bibitem{NIPS2017_7062}
S.~M. Lundberg and S.-I. Lee, ``A unified approach to interpreting model
  predictions,'' in {\em Advances in Neural Information Processing Systems 30}
  (I.~Guyon, U.~V. Luxburg, S.~Bengio, H.~Wallach, R.~Fergus, S.~Vishwanathan,
  and R.~Garnett, eds.), pp.~4765--4774, Curran Associates, Inc., 2017.

\bibitem{9373074}
A.~M.~P. Braşoveanu and R.~Andonie, ``Visualizing transformers for nlp: A
  brief survey,'' in {\em 2020 24th International Conference Information
  Visualisation (IV)}, pp.~270--279, 2020.

\bibitem{bocklisch2017rasa}
T.~Bocklisch, J.~Faulkner, N.~Pawlowski, and A.~Nichol, ``Rasa: Open source
  language understanding and dialogue management,'' {\em arXiv preprint
  arXiv:1712.05181}, 2017.

\bibitem{dlib09}
D.~E. King, ``Dlib-ml: A machine learning toolkit,'' {\em Journal of Machine
  Learning Research}, vol.~10, pp.~1755--1758, 2009.

\bibitem{li2020survey}
J.~Li, A.~Sun, J.~Han, and C.~Li, ``A survey on deep learning for named entity
  recognition,'' {\em IEEE Transactions on Knowledge and Data Engineering},
  vol.~34, no.~1, pp.~50--70, 2020.

\bibitem{chen2014fast}
D.~Chen and C.~D. Manning, ``A fast and accurate dependency parser using neural
  networks,'' in {\em Proceedings of the 2014 conference on empirical methods
  in natural language processing (EMNLP)}, pp.~740--750, 2014.

\bibitem{abney1991parsing}
S.~P. Abney, ``Parsing by chunks,'' in {\em Principle-based parsing},
  pp.~257--278, Springer, 1991.

\bibitem{zhu2018exploiting}
G.~Zhu and C.~A. Iglesias, ``Exploiting semantic similarity for named entity
  disambiguation in knowledge graphs,'' {\em Expert Systems with Applications},
  vol.~101, pp.~8--24, 2018.

\bibitem{reimers2019sentence}
N.~Reimers and I.~Gurevych, ``Sentence-bert: Sentence embeddings using siamese
  bert-networks,'' {\em arXiv preprint arXiv:1908.10084}, 2019.

\end{thebibliography}

\end{document}